\documentclass[10pt,twocolumn,letterpaper]{article}

\usepackage[pagenumbers]{iccv} 

\usepackage[table,xcdraw]{xcolor}
\usepackage{multirow}
\usepackage{tabularx}
\usepackage{float}
\usepackage{graphicx}
\usepackage{amsmath,mathrsfs}
\usepackage{amssymb}
\usepackage[colorlinks=true, allcolors=red]{hyperref}  
\usepackage{booktabs}
\usepackage{bbding}
\definecolor{iccvblue}{rgb}{0.21,0.49,0.74}

\definecolor{gray}{rgb}{0.92, 0.92, 0.92}
\definecolor{yellow}{rgb}{1, 1, 0.7}
\definecolor{orange}{rgb}{1, 0.85, 0.7}
\definecolor{red}{rgb}{1, 0.7, 0.7}

\def\paperID{4885} 
\def\confName{ICCV}
\def\confYear{2025}

\title{CityGS-\(\mathcal{X}\): A Scalable Architecture for Efficient and Geometrically Accurate Large-Scale Scene Reconstruction}

\author{Yuanyuan Gao\footnotemark[1]~~$^{1,2}$, Hao Li\footnotemark[1]~~$^{1}$, Jiaqi Chen\footnotemark[1]~~$^{1}$, Zhengyu Zou$^{1}$, \\  {Zhihang Zhong\footnotemark[2]~~$^2$}, {Dingwen Zhang\footnotemark[2]~~$^{1}$}, Xiao Sun$^{2}$, Junwei Han$^{1}$\\
$^{1}$Northwestern Polytechnical University, ~~$^2$Shanghai Artificial Intelligence Laboratory \\
}

\begin{document}
\twocolumn[{
 \renewcommand\twocolumn[1][]{#1}
\maketitle
 \thispagestyle{empty}
 \pagestyle{empty}
 \begin{center}
     \captionsetup{type=figure}
    \vspace{-2em} \includegraphics[width=1\linewidth]{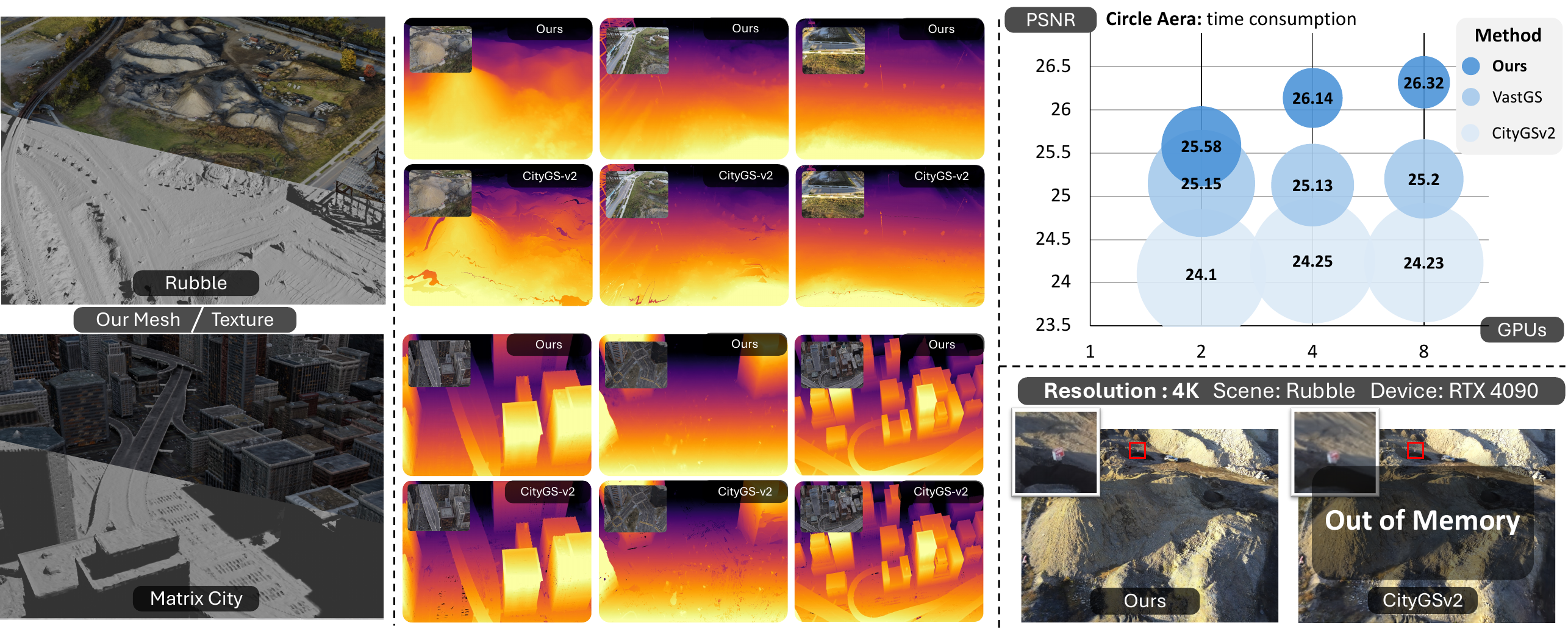}
    \vspace{-2em}
     \captionof{figure}{We propose \textbf{CityGS-\(\mathcal{X}\)}, a scalable architecture for efficient and geometrically accurate large-scale scene reconstruction (left). 
     The middle column shows qualitative results of rendered depth between our method and CityGS-v2~\cite{liu2024citygaussianv2}, demonstrating CityGS-\(\mathcal{X}\)'s superior geometric representation with smoother object surfaces. The top-right PSNR chart highlights CityGS-\(\mathcal{X}\)'s superior reconstruction quality across various GPU configurations while significantly reducing time consumption.
     The bottom-right section highlights the memory efficiency of CityGS-\(\mathcal{X}\), successfully handling high-resolution 4K rendering, while CityGS-v2 encounters out-of-memory issues.
    }
     \label{fig:overview}
 \end{center}
}]
\renewcommand{\thefootnote}{\fnsymbol{footnote}}
\footnotetext[1]{Equal Contribution.  \footnotemark[2]Corresponding author.}

\begin{abstract}
Despite its significant achievements in large-scale scene reconstruction, 3D Gaussian Splatting still faces substantial challenges, including slow processing, high computational costs, and limited geometric accuracy.
These core issues arise from its inherently unstructured design and the absence of efficient parallelization.
To overcome these challenges simultaneously, we introduce \textbf{CityGS-\(\mathcal{X}\)}, a scalable architecture built on a novel parallelized hybrid hierarchical 3D representation (PH$^2$-3D).
As an early attempt, CityGS-\(\mathcal{X}\) abandons the cumbersome merge-and-partition process and instead adopts a newly-designed batch-level multi-task rendering process. This architecture enables efficient multi-GPU rendering through dynamic Level-of-Detail voxel allocations, significantly improving scalability and performance.
To further enhance both overall quality and geometric accuracy, CityGS-\(\mathcal{X}\) presents a progressive RGB-Depth-Normal training strategy.
This approach enhances 3D consistency by jointly optimizing appearance and geometry representation through multi-view constraints and off-the-shelf depth priors within batch-level training.
Through extensive experiments, CityGS-\(\mathcal{X}\) consistently outperforms existing methods in terms of faster training times, larger rendering capacities, and more accurate geometric details in large-scale scenes. 
Notably, CityGS-\(\mathcal{X}\) can train and render a scene with 5,000+ images in just 5 hours using only 4×4090 GPUs, a task that would make other alternative methods encounter Out-Of-Memory (OOM) issues and fail completely. This implies that CityGS-\(\mathcal{X}\) is far beyond the capacity of other existing methods. \textbf{Project Page:}~{\small\color{red}\url{https://lifuguan.github.io/CityGS-X/}}.
\end{abstract}    
\section{Introduction}
\label{sec:intro}

Large-scale scene reconstruction is essential in areas such as urban planning~\cite{kuang2020real}, autonomous driving~\cite{geiger2012we, caesar2020nuscenes}, and aerial surveying~\cite{li2024dgtr}. The primary goal is to create photorealistic 3D models with accurate geometry from image collections, enabling high-quality visualization, analysis, and simulation. Recent advancements in 3D Gaussian Splatting (3DGS)~\cite{kerbl3Dgaussians} have brought this goal closer to reality, improving scene reconstruction with high-fidelity novel-view synthesis and fast rendering. Despite these advancements, current 3DGS methods still struggle with issues related to geometry accuracy, as well as inefficiencies in training and inference.

\begin{figure}
    \centering
    \includegraphics[width=1\linewidth]{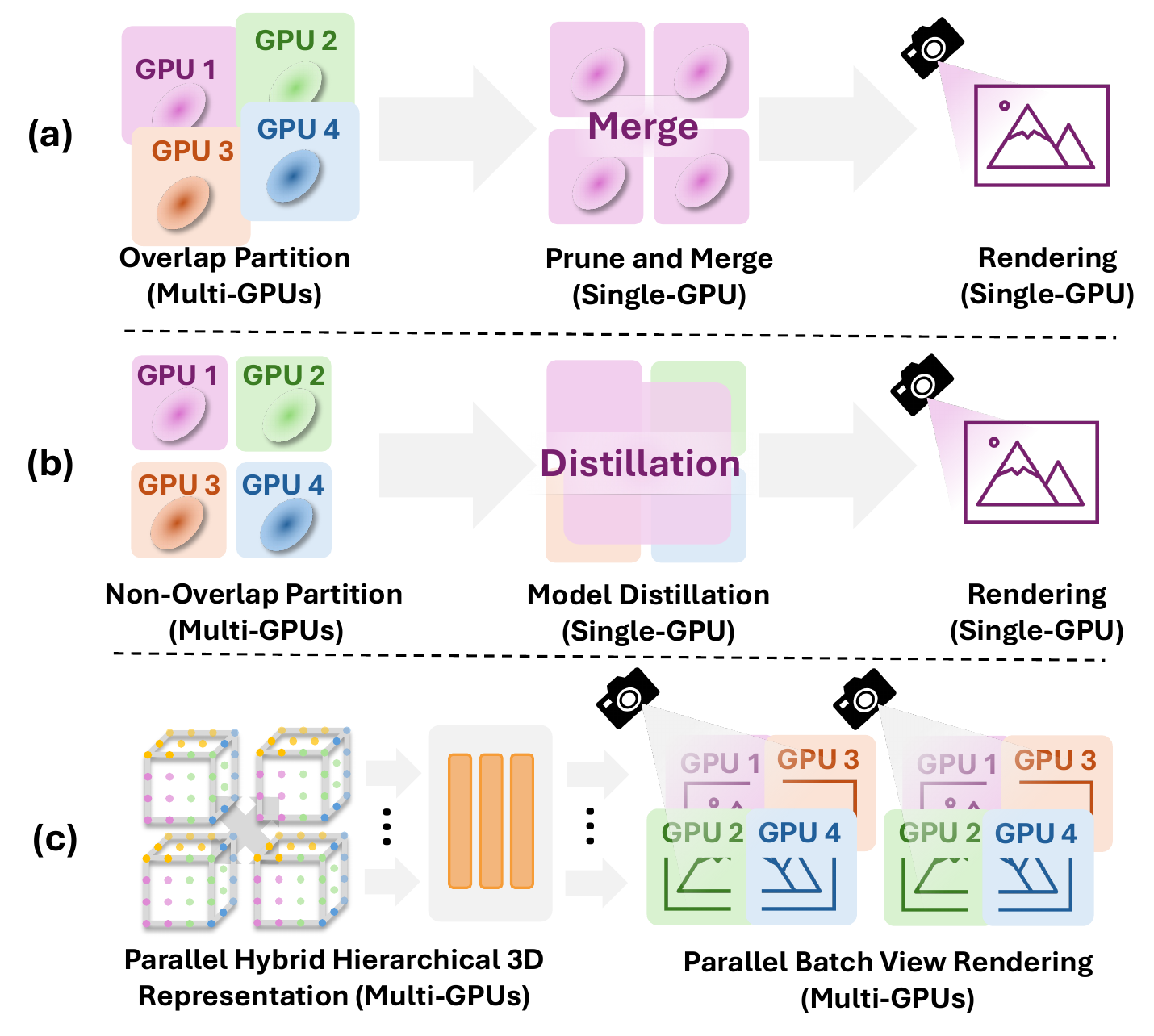}
    \caption{Comparison between our parallel architecture and previous methods that reconstruct large-scale scenes using a partition-after-merge strategy. (a) uses overlap partition training and suffers from issues in the overlap areas during merging, which is time-consuming. (b) instead employs model distillation to enable distributed training, though it still introduces extra time for the distillation process. Moreover, both (a) and (b) restrict Gaussian rendering to a single GPU, limiting the size of the final merged model. (c) Our approach introduces a novel paradigm with Parallel Hybrid Hierarchical 3D Representation and parallel batch-level rendering techniques, offering enhanced scalability and efficiency for large-scale scene reconstruction.}
    \label{fig:toutu}
    \vspace{-0.2in}
\end{figure}

Existing methods~\cite{chen2024gigags, lin2024vastgaussian, liu2024citygaussianv2} for large-scale scenes rely heavily on the merge-and-partition strategy, where scenes are divided into independent blocks for Multi-GPUs training. As shown in Fig.~\ref{fig:toutu}, these approaches face issues with inconsistencies at block boundaries due to the lack of multi-view constraints during independent block training. Some methods attempt to address this by increasing the overlap area~\cite{chen2024dogaussian, lin2024vastgaussian, liu2024citygaussianv2} or performing post-training after merging~\cite{gao2024cosurfgs, li2024dgtr}, but these solutions add extra training time and computational costs, making them inefficient for large-scale scene reconstruction. Furthermore, existing methods still require merging on a single GPU for novel-view synthesis, and even with an RTX 4090 GPU, only up to 5 million Gaussians can be handled, severely limiting scalability for large scenes.

A key insight of this work is that the field lacks a scalable architecture for efficient and geometrically accurate large-scale scene reconstruction. We argue that a compact yet geometry-friendly 3D representation, coupled with an optimized parallel training and rendering mechanism, is essential to advancing this task. Fortunately, progress has been made in both 3D representations and parallelism for 3DGS. Traditional 3DGS represents a scene using a set of discrete Gaussian attributes, but this unstructured~\cite{xld} approach fails to accurately capture the scene’s surface. It also becomes memory- and storage-intensive when handling large-scale scenes. To address these issues, hybrid representations~\cite{lu2024scaffold, ren2024octree} have been introduced, which integrate anchor or voxel structures with MLP decoders to represent Gaussians. This reduces memory usage while maintaining high-quality geometric fidelity. On the parallelism front, Grendel-GS~\cite{zhao2024scaling} is the first distributed training system designed to scale Gaussian splatting, effectively balancing the computational load across multiple GPUs.
%



Building on existing advancements, we present \textbf{CityGS-\(\mathcal{X}\)}, a novel architecture that scales up Parallelized Hybrid Hierarchical 3D Representation (PH$^2$-3D) into multi-GPU systems for scalable and efficient large-scale scene reconstruction. To achieve this goal, we dynamically allocate LOD voxel structures across multiple GPUs, enabling voxel-wise parallelism and automatic load balancing during training. Powered by a shared Gaussian decoder, the voxels across different GPUs are decoded as distributed Gaussians, facilitating parallel training and rendering. This design eliminates fragmentation, removes redundant 3D representations, and better aligns with the real geometry’s distribution. Building on this architecture, we implement batch rendering techniques with multi-task (\eg, RGB, depth, and normal), leveraging the parallel nature of the system. 
Meanwhile, since our Gaussian Decoder is an implicit representation, it is a natural fit for batch training, further enhancing its generalization ability.
More importantly, CityGS-\(\mathcal{X}\) enables the incorporation of batch-level consistent geometric constraints. We leverage the prior knowledge of an off-the-shelf mono-depth estimator~\cite{yang2024depth} for per-batch images and propose a confidence-aware filtering strategy that adaptively masks and filters low-quality depth estimations during batch-view projections, further enhancing both quality and geometric accuracy.



We perform extensive experiments on large-scale scene datasets, demonstrating the superior rendering quality, exceptional surface reconstruction performance, and fast training speed of our approach (Fig.~\ref{fig:overview}). Additionally, we test our method on various configurations of RTX 4090 GPUs, showing that it can run effectively on relatively low-end GPUs without encountering out-of-memory (OOM) issues, highlighting its scalability and efficiency.

Our key contributions are summarized as follows:
\begin{itemize}
\item We introduce \textbf{CityGS-\(\mathcal{X}\)}, a scalable architecture for efficient and geometrically accurate large-scale scene reconstruction. It leverages a Parallelized Hybrid Hierarchical 3D Representation that eliminates the merge-and-partition process.
\item We further propose multi-task batch rendering with a batch-level consistent progressive training strategy. Our approach employs robust geometric constraints at the batch level to regularize training by filtering out uncertain depth estimates across multiple views.
\item Extensive experiments demonstrate that our method outperforms existing large-scale scene reconstruction techniques, establishing a promising foundation for future advancements in the field.
\end{itemize}
    
\section{Related works}
\subsection{Neural Rendering}
Traditional 3D reconstruction methods primarily rely on Structure-from-Motion (SfM)~\cite{schonberger2016structure} and Multi-View Stereo (MVS)\cite{goesele2007multi}, representing the scenes under the format of point clouds~\cite{lhuillier2005quasi}, volumes~\cite{kutulakos2000theory}, or depth maps~\cite{schoenberger2016mvs, campbell2008using}.
However, such approaches often suffer from artifacts due to erroneous matching and noise. 
Recently emerged implici and explicit representations, such as NeRF~\cite{mildenhall2021nerf, Ye2023IntrinsicNeRF, ming2022idf, li2024gpnerf} and 3DGS~\cite{kerbl3Dgaussians, li2024ggrt, yan2024multi,ren2024octree,lu2024scaffold}, show promising performance in providing high-fidelity novel-view synthesis and high-speed rendering.
Nevertheless, these methods are mainly designed for photorealistic rendering and thus struggle with capturing accurate geometry and physical surfaces. 
In response to these challenges, recent advancements in Gaussian Splatting methods~\cite{chen2024pgsr, guedon2023sugar, huang20242d, tang2024hisplat} have focused on decomposing 3D Gaussian shapes into more simplified components.
However, existing methods overlook computational efficiency and resource consumption, resulting in long training times for large-scale scenes and potential out-of-memory issues with limited resources.

\subsection{Large-scale Scene Reconstruction}
Achieving photorealistic and geometrically precise reconstructions of large-scale real-world scenes has wide-ranging applications in virtual reality~\cite{turki2022mega}, autonomous driving~\cite{caesar2020nuscenes}, and urban planning~\cite{li2023matrixcity}.
%
%
%
Recently, 3DGS~\cite{kerbl3Dgaussians} has emerged as a promising alternative, offering real-time rendering with high visual fidelity. Moreover, Octree-GS~\cite{ren2024octree} combines a hybrid representation~\cite{lu2024scaffold} into a level-of-detail (LoD) structure to reduce redundant Gaussian points without sacrificing performance.
%
%
Meanwhile, many methods~\cite{lin2024vastgaussian,chen2024dogaussian,liu2024citygaussian} like VastGS~\cite{lin2024vastgaussian}  employ a partition-and-merge strategy that divides the entire scene into multiple areas for efficient training convergence. 
However, they primarily focus on novel-view synthesis and lack interaction between blocks, leading to inconsistencies at block boundaries due to the absence of multi-view constraints.
Moreover, some works~\cite{liu2024citygaussianv2,gao2024cosurfgs,chen2024gigags} integrate geometric optimized methods like PGSR~\cite{chen2024pgsr} or 2D-GS~\cite{huang20242d} into large scene reconstruction.
Nevertheless, the methods mentioned above rely on novel-view rendering on a single GPU, which necessitates merging all blocks onto a single device, a process that is time-consuming. Although several pruning methods exist~\cite{fan2024lightgaussian,liu2024maskgaussian}, while effective to some degree, the memory capacity remains constrained by a single GPU's limitations. This limitation hampers their scalability for large-scale scene reconstruction due to the memory constraints of a single GPU.
In contrast, our method scales up parallelism hybrid 3D representation into parallel training and rendering tasks and further proposes a batch-level geometric constraint technique, which successfully addresses the challenges of scalability, efficiency, and accuracy.
\begin{figure*}[!t]
    \centering
    \includegraphics[width=1\linewidth]{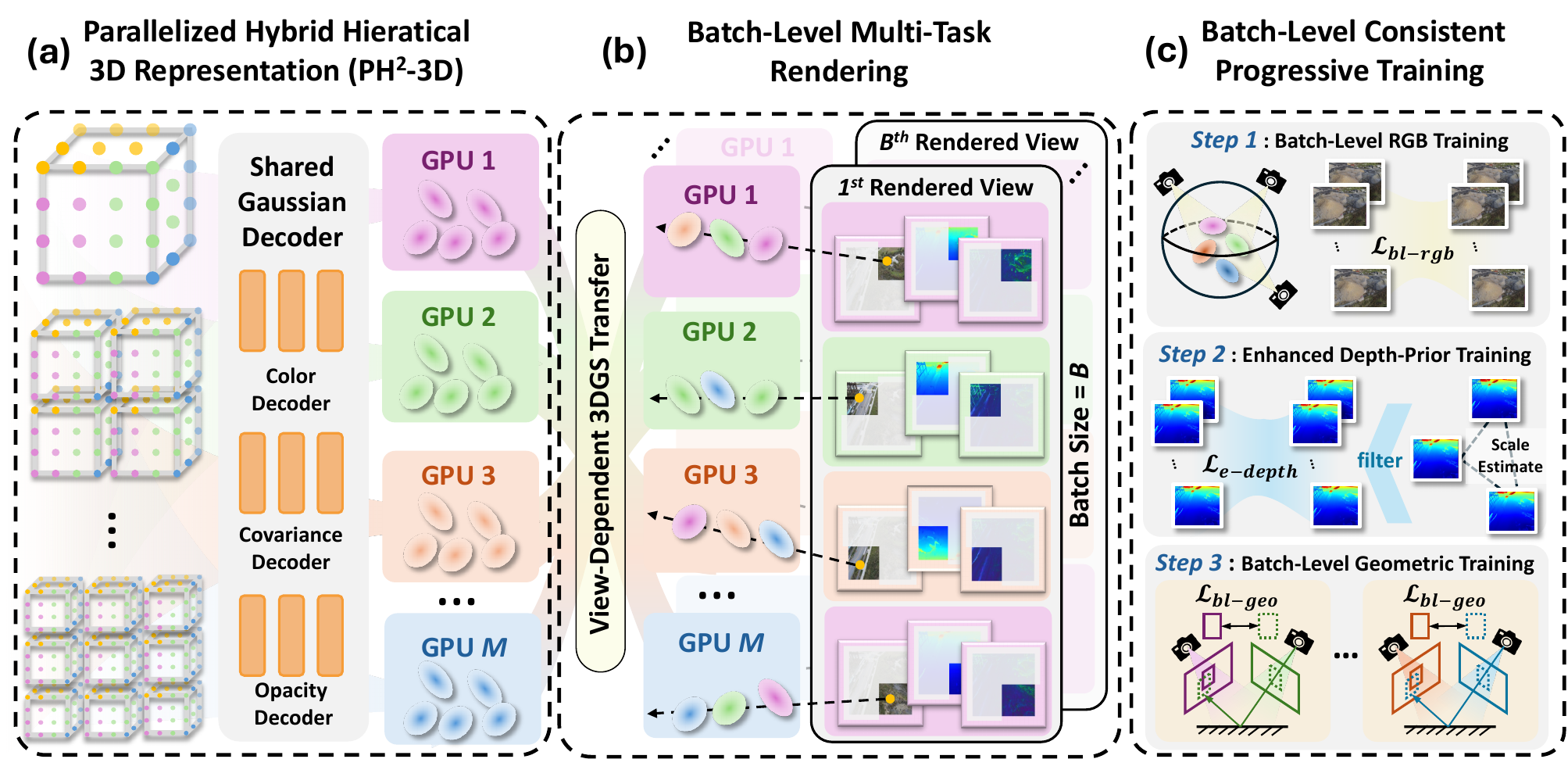}
    \caption{\textbf{CityGS-\(\mathcal{X}\)} is a scalable framework that eliminates the partition-and-merge paradigm by utilizing a parallel training as well as rendering technique. (a) PH\(^2\)-3D dynamically allocates LoD voxels across multiple GPUs for steps in training (Sec.~\ref{sec:representation}). (b) The distributed multi-task rendering strategy divides images to be rendered into patches and assigns them to different GPUs for rendering RGB/Depth/Normal in parallel (Sec.~\ref{sec:rendering}). (c) Building upon this framework, we introduce a novel progressive training strategy that applies multi-view consistency on batch training for appearance quality and geometry accuracy (Sec.~\ref{sec:training}). }
    \label{fig:framework}
    \vspace{-0.2in}
\end{figure*}

\section{Method}
\subsection{Preliminaries}
\noindent \textbf{3DGS~\cite{kerbl3Dgaussians}.} 3D gaussian splatting explicitly represents 3D scenes with anisotropic Gaussian primitives. Each Gaussian is parameterized by  position \( x \), mean \( \mu \),  covariance matrix \(\Sigma\), opacity \(\sigma\) , and color \(c\). During rendering, it is effectively rendered by computing the color $C$ of a pixel by blending $N$ ordered Gaussians overlapping the pixel:
\begin{equation}
\boldsymbol{C}=\sum_{i \in N} \boldsymbol{c}_i \alpha_i \prod_{j=1}^{i-1}\left(1-\alpha_j\right),
\end{equation}
where \(\alpha_i\) is evaluating a Gaussian with its covariance \(\sum_i\) multiplied with its opacity \(\sigma_i\). This design provides differentiable rendering in real time.

\noindent \textbf{Octree-GS~\cite{ren2024octree}.} The octree structure in Octree-GS organizes the scene into a hierarchical grid, where each level of the octree represents a different scale of the scene. The Level-of-Detail (LOD) for each region is dynamically selected based on the observation distance and scene complexity. 
This octree structure significantly reduces the number of primitives required for rendering by ensuring that only the relevant Gaussians from appropriate LOD levels are selected, thus enabling real-time performance even in complicated scenes with millions of primitives. 

\subsection{Parallelized Hierarchical Hybrid 3D 
}
\label{sec:representation}
Given the inherent difficulties in achieving real-time rendering across various scales,
especially for conventional 3DGS methods, the surface reconstruction of large-scale scenes has become a challenging task.

To eliminate partition-and-merge and achieve scalable training, we design our representation as DDP-like~\footnote{DistributedDataParallel (DDP) is a module designed for parallel training of models across multiple GPUs or nodes} paradigm, as shown in Fig.~\ref{fig:framework}. 
Specifically, we adopt a \(K\)-LoDs hierarchical hybrid structure $\left\{\mathbf{X}_k\right\}_{k=1}^K$ to represent the large-scale scene, where each LoD consists of different sizes voxel sets $\mathbf{X}_k=\left\{X_{k,1}, X_{k,2}, \ldots, X_{k,v}\right\}$, \(v\) is the total number of the voxels. The formulation is shown below:
\begin{equation}
\mathbf{X}_k=\left\{\left\lfloor\frac{\mathbf{P}}{\delta / 2^k}\right\rceil \cdot \delta / 2^k\right\}, \quad 0 \leq k \leq K-1,
\end{equation}
where \(\delta\) denotes the initial voxel size on 0-th LoD, \(\lfloor\cdot\rceil\) denotes the rounding operations, and \(\mathbf{P}\) is the sparse point clouds initialized by SfM.
To enable parallel computation, we distribute the voxels \(\mathbf{X}_k\) of different LoDs across different GPUs. However, load imbalance among different GPUs is a performance bottleneck in parallel training because GPUs assigned fewer computational tasks must wait for other GPUs to complete their tasks.
To this end, we adopt a spatial average sampling strategy to ensure that the number of voxels activated on each GPU during view-dependent rendering is equal:
\begin{equation}
\begin{cases}\mathbf{X}_{k}^{(m)} =\left\{ X_{k,i}^{(m)}\mid i=m+jM \right\}\\ j\leq \frac{v}{M} ,\quad m\in \left\{ 1,\cdots ,M \right\}\end{cases},
\end{equation}
where \(M\) denotes the number of GPUs, \(\mathbf{X}_k^{(m)}\) is the assigned voxels on \(m\)-th GPU. Such a strategy maximizes the overall parallel processing capability, speeding up the overall training task.

For each voxel \(X^{(m)}_{k,v}\), we assign a learnable embedding \(F^{(m)}_v\in\mathbb{R}^{32}\), 
a scaling factor $l^{(m)}_v \in \mathbb{R}^3$, position \(x^{(m)}_v\), and $n$ learnable offsets $O^{(m)}_v \in \mathbb{R}^{n \times 3}$. 
To transfer voxels into Gaussians, we distribute a set of shared Gaussian Decoders \(\mathrm{F}_{de}(\cdot)\) across multiple GPUs, allowing them to access global information through multi-GPUs gradient synchronization.
Then voxels pass through the Decoder \(\mathrm{F}_{de}(\cdot)\) to predict \(n\) Gaussian attributes in parallel manner:
 \begin{equation}
 \label{eq:render}
 \begin{array}{rl}
      \{\mu^{(m)}_n\}&=\{x^{(m)}_v\}+O^{(m)}_n\cdot l^{(m)}_v,\\
      \left\{\{\alpha^{(m)}_n\},\{c^{(m)}_n\},\{\Sigma^{(m)}_ n\}\right\}&=\mathrm{F}_{de}\left(F^{(m)}_v, \Delta_{\mathrm{vc}}, \tilde{\mathrm{~d}}_{\mathrm{vc}}\right), \\
 \end{array}
\end{equation}
where \(\Delta_{\mathrm{vc}}\) and \( \tilde{\mathrm{~d}}_{\mathrm{vc}}\) denote the relative viewing distance and the direction to the camera.  
This structure significantly reduces the memory cost by storing explicit Gaussian attributes. It also reduces the inter-GPU communications, as synchronization is required only for the decoder's gradients.
%

\subsection{Batch-Level Multi-Task Rendering}
\label{sec:rendering}
Traditional large-scale scene reconstruction methods partition areas into several blocks in spital and train these blocks through different GPU independently. 
It forces them to merge all blocks into a single device before performing novel-view synthesis, further limiting their scalability on large-scale scene rendering since the memory of single GPU is limited.
In contrast, benefiting from our parallelism representation, it is a natural fit for us to present parallel rendering techniques by directly rendering images from multiple GPUs, enabling efficient and large-scale scene rendering with larger scales or resolutions.

As shown in Fig.~\ref{fig:framework}, CityGS-\(\mathcal{X}\) distributes the image rendering tasks areas across GPUs for the multi-task rendering and loss computation. In practice, we patchify the batch-rendered images into serialized \(16\times 16\)-pixel patches and then distribute the patches to different GPUs using an adaptive load-balancing strategy following Grendal-GS~\cite{zhao2024scaling}. 
During each patch rendering task, we propose View-Dependent Gaussian Transfer strategy, which searches the intersected voxels across all LoDs on all GPUs in parallel, predicts their corresponding Gaussian attributes (as Eq.~\ref{eq:render}), and then transfers them into the GPU assigned to this task.
After that, all blocks perform parallel rendering following classical tile-based rasterization:
\vspace{-0.05in}
\begin{equation}
\hat{\boldsymbol{C}}=\sum_{i=1}^N \boldsymbol{c}_{\pi(i)} \alpha_{\pi(i)} \prod_{j=1}^{i-1}\left(1-\alpha_{\pi(j)}\right),
\vspace{-0.05in}
\end{equation}
where \(N\) denotes the number of transferred Gaussian points and \(\pi(\cdot)\) is the reordered function for transferred Gaussian.
Additionally, we obtain the normal map of the current viewpoint following the strategy of PGSR~\cite{chen2024pgsr}:
\begin{equation}
\hat{\boldsymbol{N}}=\sum_{i=1}^N  R_c^T n_{\pi(i)} \alpha_{\pi(i)} \prod_{j=1}^{i-1} (1 - \alpha_{\pi(j)}),
\end{equation}
where \( R_c^T \) is the rotation matrix from camera to world coordinates and \( n_{\pi(i)} \) is the normal of \( \pi(i) \)-th reordered Gaussians.
Moreover, we compute depth maps using unbiased depth rendering technique~\cite{chen2024pgsr}, which considers depth as the intersection between rays and the Gaussian plane:
\begin{equation}
\begin{cases}
\displaystyle
\boldsymbol{D} = \sum_{i=1}^N d_{\pi(i)}\, \alpha_{\pi(i)}\, \prod_{j=1}^{i-1} \Bigl(1 - \alpha_{\pi(j)}\Bigr), \\[10pt]
\hat{\boldsymbol{D}} = \dfrac{\boldsymbol{D}}{\hat{\boldsymbol{N}}\, \boldsymbol{K}^{-1}\, \tilde{\boldsymbol{p}}},
\end{cases}
\end{equation}
where \(\boldsymbol{D}\) denotes the distance map through alpha-blending, \( d_{\pi{i}} \) denotes the distance of \( \pi(i) \)-th reordered Gaussians. \(\tilde{\boldsymbol{p}}\) is the homogeneous coordinate representation and \(\boldsymbol{K}^{-1}\) is the inverse of camera intrinsics.
Notably, traditional methods, such as MVGS~\cite{du2024mvgs}, rely on gradient accumulation are limited by the memory constraints of a single GPU.
Since we parallelize our whole training into multiple GPUs, it enables us to scale up training batches flexibly (up to 32 in our experiments).

\begin{figure}[t]
\centering\includegraphics[width=1\linewidth]{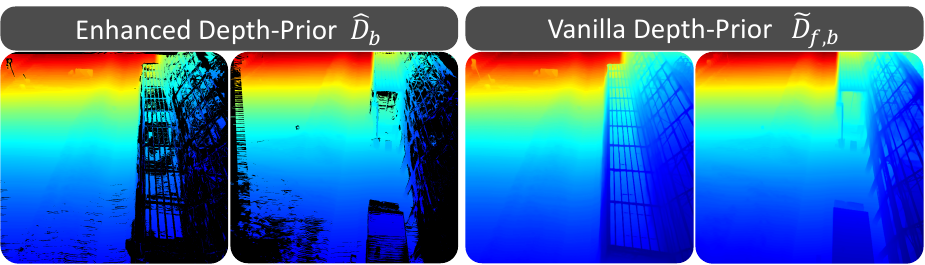}
    \caption{Visualization of enhanced and vanilla pseudo depth. In enhanced depth, multi-view inconsistency depth information is filtered out and shown as black regions. }
    \label{fig:e-depth}
\end{figure}

\begin{table*}[t]
 \centering
   \caption{\textbf{Quantitative results of novel view synthesis on Mill19~\cite{turki2022mega} dataset 
          and UrbanScene3D~\cite{lin2022capturing} dataset}. $\uparrow$: higher is better, $\downarrow$: lower is better.
          The \colorbox{red}{red}, \colorbox{orange}{orange}, and \colorbox{yellow}{yellow} colors respectively 
          denote the best, the second best, and the third best results.
          \textbf{Bold} denotes the best result in the `With Geometric Optimization' group. $\dagger$ denotes without applying the decoupled appearance encoding.
 }
    \resizebox{\linewidth}{!}{
        \LARGE
        \begin{tabular}{l ccc ccc ccc ccc}
            \toprule[1.1pt]
             &   \multicolumn{3}{c}{\emph{Building}}  
             &   \multicolumn{3}{c}{\emph{Rubble}} 
             &  \multicolumn{3}{c}{\emph{Residence}} 
             &   \multicolumn{3}{c}{\emph{Sci-Art}} \\
             \cmidrule(r){2-4} \cmidrule(r){5-7} \cmidrule(r){8-10} \cmidrule(r){11-13} 
            &  SSIM$\uparrow$ & PSNR$\uparrow$ & LPIPS$\downarrow$   
            &  SSIM$\uparrow$ & PSNR$\uparrow$ & LPIPS$\downarrow$ 
            &  SSIM$\uparrow$ & PSNR$\uparrow$ & LPIPS$\downarrow$   
            &  SSIM$\uparrow$ & PSNR$\uparrow$ & LPIPS$\downarrow$  \\
            \midrule
            \textbf{w/o Geometric Optimization}  \\
            \midrule

            Mega-NeRF    
             & 0.547 & 20.92 & 0.454 
             & 0.553 & 24.06 & 0.508 
             & 0.628 & 22.08 & 0.401 
             & 0.770 & \cellcolor{orange}{25.60} & 0.312 \\
            
            Switch-NeRF  
             & 0.579 & 21.54 & 0.397 
             & 0.562 & 24.31 & 0.478
             & 0.654 & \cellcolor{red}{22.57} & 0.352 
             & 0.795 & \cellcolor{red}{26.51} & 0.271 \\

            VastGaussian $\dagger$
            & 0.728 & 21.80  & 0.225 
            & 0.742 & 25.20  & 0.264 
            & 0.699 & 21.01 & 0.261 
            & 0.761 & 22.64 & 0.261 \\

            3DGS
            & 0.738 & 22.53 & 0.214 
            & 0.725 & 25.51 & 0.316 
            & 0.745 & \cellcolor{yellow}{22.36} & 0.247 
            & 0.791 & 24.13 & 0.262 \\ 

            DoGaussian
            & 0.759 & \cellcolor{yellow}{22.73} & \cellcolor{yellow}{0.204}
            & 0.765 & \cellcolor{yellow}{25.78} & 0.257 
            & 0.740 & 21.94 & 0.244
            & 0.804 & \cellcolor{yellow}{24.42} & \cellcolor{yellow}{0.219} \\

            Momentum-GS  
            & \cellcolor{orange}{0.815} & \cellcolor{red}{23.23} & \cellcolor{orange}{0.194} 
            & \cellcolor{red}{0.827} & \cellcolor{orange}{25.93} & \cellcolor{red}{0.201} 
            & \cellcolor{orange}{0.818} & 22.21 & \cellcolor{orange}{0.197} 
            & \cellcolor{orange}{0.856} & 23.02 & \cellcolor{orange}{0.205} \\
            
            CityGaussian  
            & \cellcolor{yellow}{0.778} & 21.55 & 0.246 
            & \cellcolor{yellow}{0.813} & 25.77 & \cellcolor{yellow}{0.228}
            & \cellcolor{yellow}{0.813} & 22.00 & \cellcolor{yellow}{0.211}
            & \cellcolor{yellow}{0.837} & 21.39 & 0.230 \\

            \midrule

            \textbf{w/ Geometric Optimization}  \\
            \midrule
            SuGaR   
            & 0.507 & 17.76 & 0.455 
            & 0.577 & 20.69 & 0.453 
            & 0.603 & 18.74 & 0.406 
            & 0.698 & 18.60 & 0.349 \\

            NeuS     
             & 0.463 & 18.01 & 0.611 
             & 0.480 & 20.46 & 0.618 
             & 0.503 & 17.85 & 0.533 
             & 0.633 & 18.62 & 0.472 \\

             Neuralangelo & 0.582 & 17.89 & 0.322
             & 0.625 & 20.18 & 0.314
             & 0.644 & 18.03 & 0.263
             & 0.769 & 19.10 & 0.231 \\
             
            PGSR
            & 0.480 & 16.12 & 0.573 
            & 0.728 & 23.09 & 0.334 
            & 0.746 & 20.57 & 0.289 
            & 0.799 & 19.72 & 0.275 \\
            PGSR+VastGS
            & 0.720 & 21.63 & 0.300
            & 0.768 & 25.32 & 0.274
             & -- & -- & --
             & -- & -- & -- \\


            CityGaussianV2  
            & 0.650 & 19.07 & 0.397 
            & 0.720 & 23.75 & 0.322 
            & 0.769 & 21.15 & 0.234
            & 0.810 & 20.66 & 0.266 \\
            \midrule
            \textbf{CityGS-\(\mathcal{X}\) (Ours)}  
            & \cellcolor{red}{\textbf{0.817}} & \cellcolor{orange}{\textbf{22.76}} & \cellcolor{red}{\textbf{0.191}} 
            & \cellcolor{orange}{\textbf{0.823}} & \cellcolor{red}{\textbf{26.15}} & \cellcolor{orange}{\textbf{0.210}}
            & \textbf{\cellcolor{red}{0.819}} & \textbf{\cellcolor{orange}{22.44}} & \textbf{\cellcolor{red}{0.194}} 
            & \textbf{\cellcolor{red}{0.867}} &\textbf{22.77} & \textbf{\cellcolor{red}{0.179}}
            \\

            \bottomrule[1.1pt]
        \end{tabular}
    }
    \label{tab: compare}
\centering
\end{table*}

\begin{figure*}[!h]
    \centering
    \includegraphics[width=1.0\linewidth]{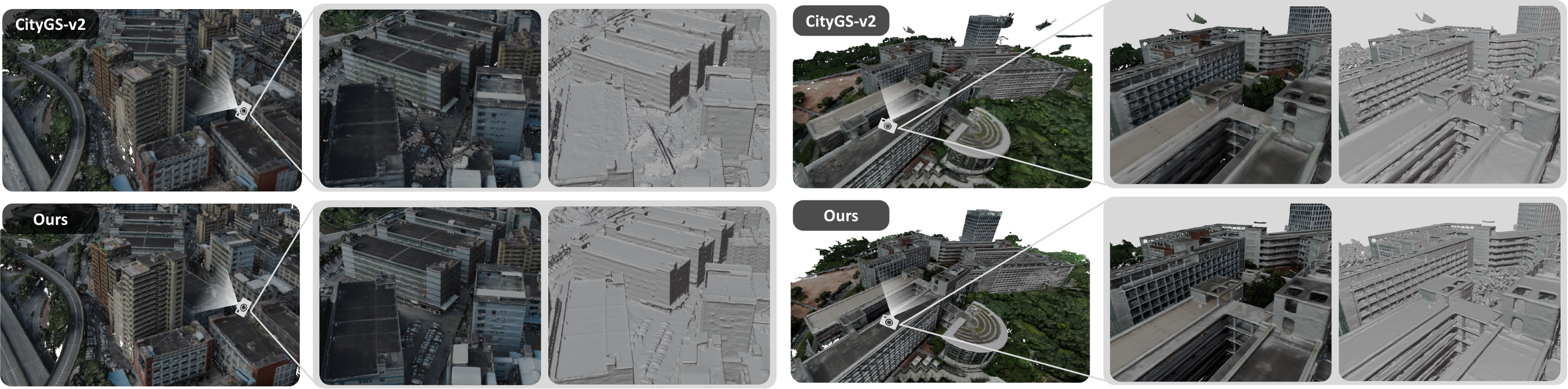}
    \caption{Qualitative mesh and texture comparison between CityGS-v2~\cite{liu2024citygaussianv2} and our method on the Residence and Sci-Art scenes~\cite{lin2022capturing}.}
    \label{fig:mesh_vis}
    \vspace{-0.25in}
\end{figure*}

\subsection{Batch-Level Consistent Progressive Training}
\label{sec:training}
To further boost the appearance quality and geometric accuracy of our proposed methods, we apply the multi-view consistency of our batch-level rendering and propose a progressive training technique, which divides into three steps: (1) a batch-level RGB training to improve the generalization of our Gaussian Decoder; (2) Enhanced Depth-Prior Training to reconstruct smoother and more accurate surface of the scene; (3) Batch-Level Geometric Training to refine the detailed of both appearance and geometry of the scene with geometric rule-based supervision.

\noindent \textbf{Step 1: Batch-Level RGB Training.} It is well known that a suitable batch size boosts the neural network's performance~\cite{yao2018large}. Since our method performs multiple images rendering in a batch, it is a natural fit for us to scale up the regular single-view RGB training to batch-level \(\mathcal{L}_{bl-rgb}\):
\begin{equation}
\mathcal{L}_{bl-rgb}=\frac{1}{B} \sum_{b=1}^B\left\|\hat{I}_b-I_b\right\|_1,
\label{eq:mv-rgb}
\end{equation}
where \(B\) is the predefined batch size, \(\hat{I}_b\) and \(I_b\) are the rendered and GT RGB images separately.
Such a strategy enables the shared Gaussian Decoder to update gradients across multi-view images, helping to mitigate the overfitting problem toward specific viewpoints.

\noindent \textbf{Step 2: Enhanced Depth-Prior Training.} Recently emerged off-the-shelf monocular depth estimator methods~\cite{yang2024depth} have smooth and continuous object surfaces as well as clear object boundaries, which can serve as guidance for depth regularization. However, it is non-trivial to apply these priors since such estimations lack the actual scale of the pseudo depth and often suffer from poor multi-view consistency in long-range areas~\cite{wang2024shape}. 
To this end, we adopt the least squares algorithm to recover the pseudo depth \(\tilde{\boldsymbol{D}}_p\) with real scale by computing the scale and shift factors between pseudo depth \(\boldsymbol{D}_p\) and sparse point clouds \(\mathbf{P}\). 
After that, we propose an Enhanced Depth-Prior Regularization \(\mathcal{L}_{e-depth}\), where we filter out multi-view inconsistency region on the predicted depth \(\tilde{\boldsymbol{D}}_p\) before applying L1 depth regularization. Specifically, we compute re-projection error \(\boldsymbol{E}\) between the target view and nearby views, then we obtain enhanced pseudo depth \(\tilde{\boldsymbol{D}}_f\) filter the region with large errors using the threshold \(\tau_{d}\), as shown in Fig.~\ref{fig:e-depth}.
During training, we directly regularize the rendered depth \(\hat{\boldsymbol{D}}\) with filtered pseudo depth \(\tilde{\boldsymbol{D}}_f\) in a batch similar to Eq.~\ref{eq:mv-rgb}:
\begin{equation}
\mathcal{L}_{e-depth}= \frac{1}{B} \sum_{b=1}^B\left\|\hat{\boldsymbol{D}_b}-\tilde{\boldsymbol{D}}_{f,b}\right\|_1 .
\end{equation}
Such a mechanism enables us to produce a smooth surface of the scene and remove artifacts from the scene, reducing the difficulties for further geometric learning.

\noindent \textbf{Step 3: Batch-Level Geometric Training.} Although step 2 introduces a powerful prior for depth regularization, its prediction is still imperfect in small objects. Thus, we introduce batch-level geometric-based constraints for refinement. 
Particularly, we divide the rendered bach images into pairs \(\{(\hat{\boldsymbol{C}}_1,\hat{\boldsymbol{C}}_2),\cdots,(\hat{\boldsymbol{C}}_{2i-1},\hat{\boldsymbol{C}}_{2i})\}\). 
For each pair \((\hat{\boldsymbol{C}}_{2i-1},\hat{\boldsymbol{C}}_{2i})\), we randomly map a \(7\times 7\) pixel patch centered at \(\boldsymbol{p}_{2n}\) to the pair image patch  using the homography matrix \(\boldsymbol{H}_{2n-1,2n}\).  
Then, we apply multi-view photometric constraints by minimizing the reprojection errors between the patches of batch views. The overall formulation is shown below:
\begin{equation}
\small
\begin{split}
\mathcal{L}_{bl-geo} = \frac{1}{B/2} \sum_{i=0}^{B/2} \Biggl( 1 - \operatorname{NCC}\Bigl(
\hat{\boldsymbol{C}}_{2i}\left(\boldsymbol{p}_{2i}\right), \\
\quad \text{sg}\Bigl[\hat{\boldsymbol{C}}_{2i-1}\left(\boldsymbol{H}_{i,2i-1}\boldsymbol{p}_{2i}\right)\Bigr]
\Bigr) \Biggr),
\end{split}
\end{equation}
%
where sg[$\cdot$] denotes stop gradient, NCC($\cdot$) denotes normalized cross correlation~\cite{yoo2009fast}.
\begin{figure*}[t]
    \centering
    \includegraphics[width=1\linewidth]{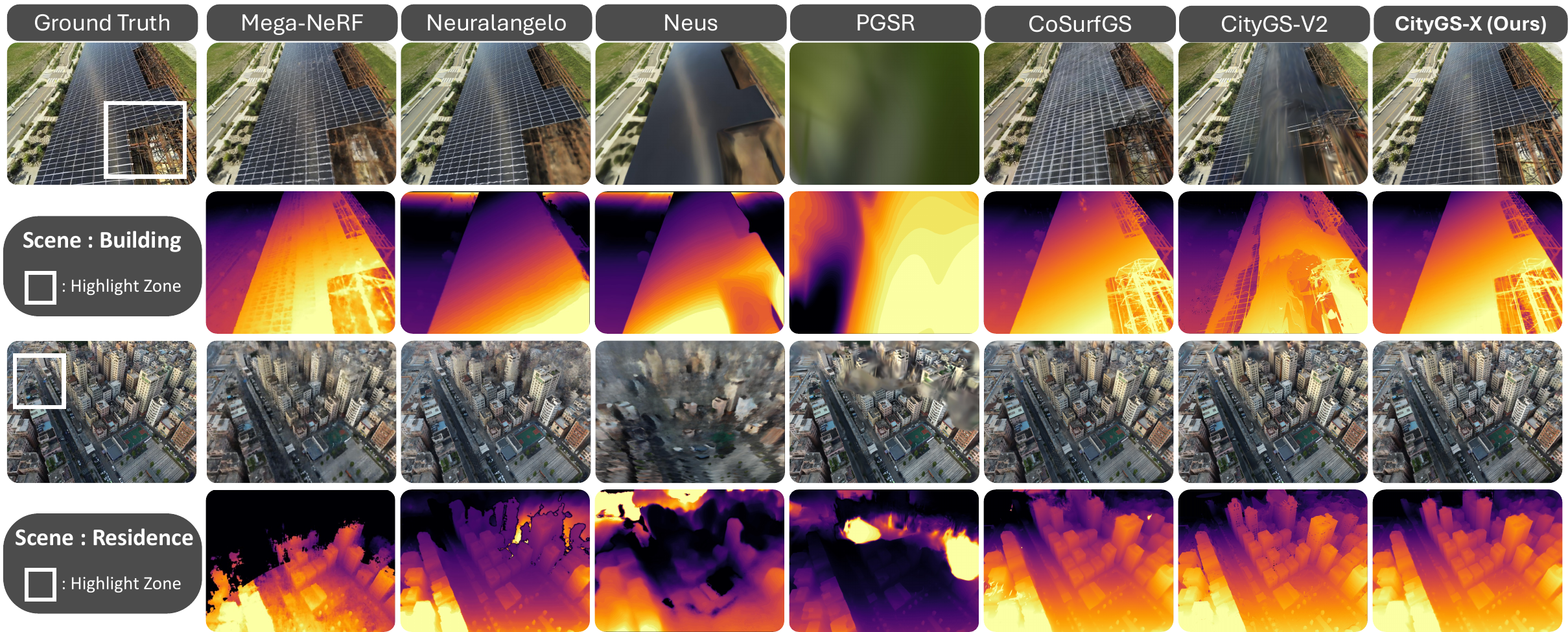}
    \caption{Qualitative results of ours and other methods in image and depth rendering on Mill-19~\cite{turki2022mega} and Urbanscene3D~\cite{{turki2022mega}} datasets.}
    \label{fig:rgb_depth_vis}
\end{figure*}

\begin{figure*}[t]
    \centering
    \includegraphics[width=1\linewidth]{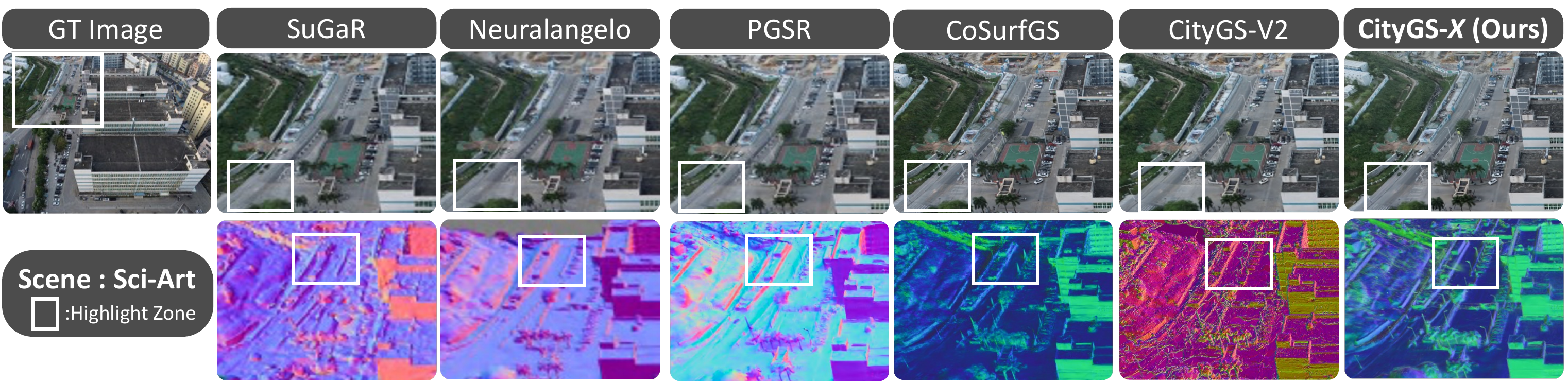}
    \caption{Qualitative results of our method and other methods in large-scale reconstruction datasets  UrbanScene~\cite{lin2022capturing}. Here we demonstrate the normal visualization and the corresponding rendered RGB images. The discriminate areas are zoomed up by `$\Box$'.}
    \label{fig:normal_visual}
\end{figure*}

\section{Experiments}
\subsection{Experimental Setup}
We conduct experiments on five scenes from three datasets: Building and Rubble from the Mill-19~\cite{turki2022mega}, Residence and Sci-Art from UrbanScene3D~\cite{lin2022capturing}, and Aerial from a small city region within the MatrixCity~\cite{li2023matrixcity} dataset. All experiments are performed on \(4\times\)RTX 4090 GPUs with a batch size of \(B=4\), unless stated otherwise. 
See additional details on the evaluation and training process in the supplementary materials.

\subsection{Main Results}
\noindent \textbf{Novel View Synthesis.} In Tab.~\ref{tab: compare} and Fig.~\ref{fig:rgb_depth_vis}, we conduct both quantitative and qualitative comparisons to evaluate the rendering quality of recent large-scale scene reconstruction methods \textit{w/} and \textit{w/o} geometric optimizations.
It is evident that CityGS-\(\mathcal{X}\) achieves the state-of-the-art even compared with the methods without geometry constraints, \textit{i.e}, a 0.2dB improvement in terms of PSNR on the Rubble~\cite{turki2022mega} and a 0.026 reduction in terms of LPIPS on the Sci-Art~\cite{lin2022capturing}. 
Qualitative results in Fig.~\ref{fig:rgb_depth_vis} demonstrate that our method effectively captures fine-grained geometric details while avoiding floating artifacts.
This highlights its strong ability to maintain multi-view consistency and preserve appearance information across a vast collection of images, ultimately enabling large-scale scene reconstruction with highly precise geometric representation.

\noindent \textbf{Surface Reconstruction.} 
In this part, we compare our method with other surface reconstruction methods on the MatrixCity~\cite{li2023matrixcity}. The experimental results in Table~\ref{tab: compare_reconstruction} demonstrate that our method achieves state-of-the-art performance with a PSNR of 27.58 dB, surpassing CityGS-V2~\cite{liu2024citygaussianv2} by a significant margin of 0.35 dB across all evaluation metrics.
Fig.~\ref{fig:normal_visual} presents a comparison between our method and other reconstruction approaches on the UrbanScene dataset. 
Our method achieves sharper details, better geometry, and smoother reconstructions, outperforming others in visual quality.
Additionally, Fig.~\ref{fig:mesh_vis} presents qualitative comparisons between our method and CityGS-V2 in different scenes. It can be observed that our method produces more detailed and clearer surface structures, closely resembling real-world urban environments. 
The experiment above demonstrates that our method effectively learns geometric representations, achieving precise surface reconstruction.

\begin{figure}[t]
    \centering    \includegraphics[width=1\linewidth]{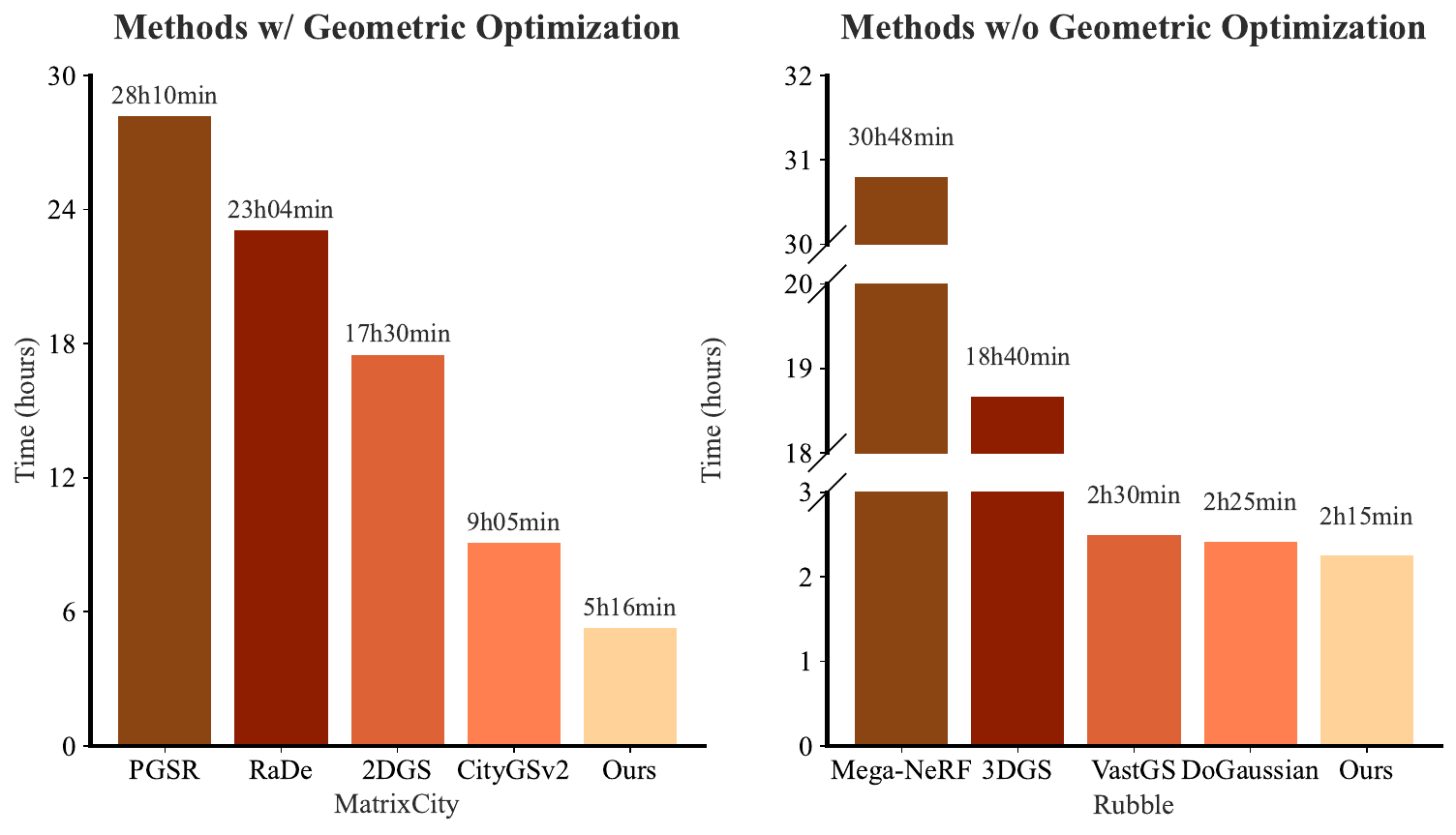}
    \caption{Time consumptions between our method and others.}
    \label{fig:time_consumption}
\end{figure}

\noindent \textbf{Training Time Consumption.}
We compare the training times of our method with existing approaches, categorized into methods \textit{w/} and \textit{w/o} geometric optimization. The results are shown in Fig.~\ref{fig:time_consumption}. For methods with geometric optimization (left), our approach completes training in just 5 hours, significantly faster than other methods. For methods without geometric optimization (right), our method finishes training in only 2 hours and 15 minutes, even slightly outperforming methods that do not incorporate geometric considerations. These results highlight the efficiency and scalability of our approach, making it well-suited for large-scale scene reconstruction.


\subsection{Ablation Studies}

\begin{figure}
    \centering\includegraphics[width=0.9\linewidth]{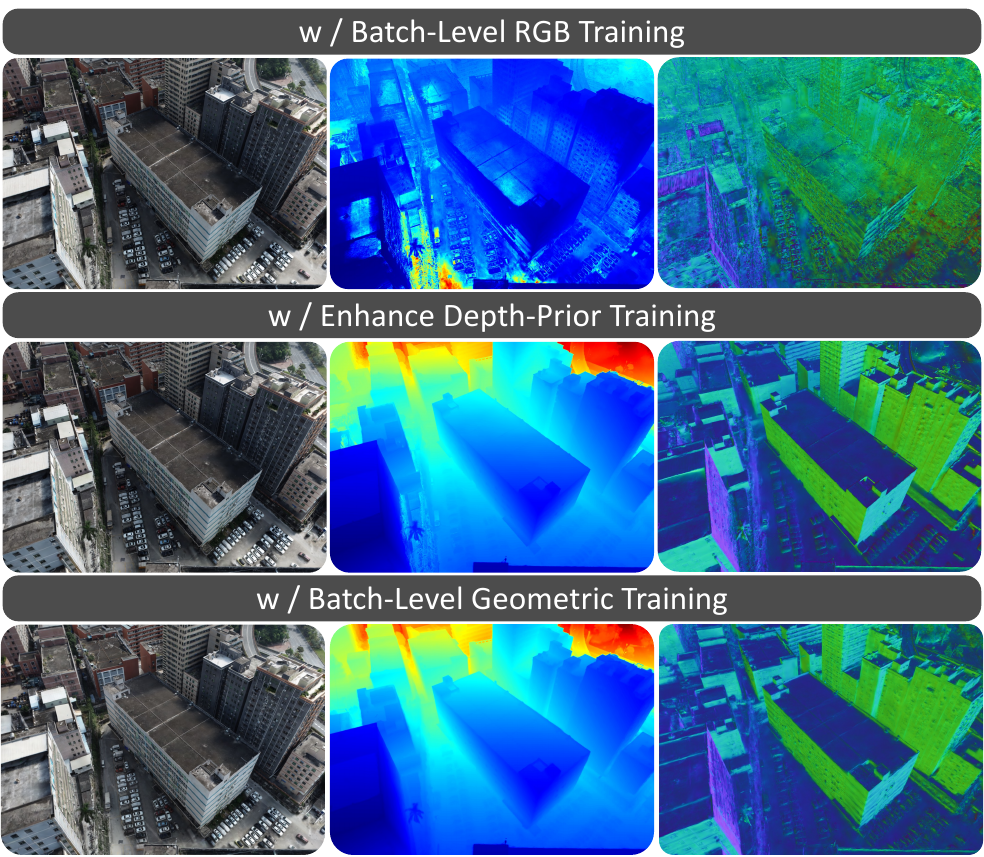}
    \vspace{-0.1in}
    \caption{Progressive Training Strategy on Residence.}
    \label{fig:ablation_vis}
    \vspace{-0.2in}
\end{figure}

\noindent \textbf{Progressive Training Strategy}.
We conduct ablation experiments on both Building and Residence datasets. The results are reported in Tab.~\ref{tab:progressive}.  
Our Batch-Level RGB Training in ID. 2 reduces training time by nearly 3 times and achieves an improvement of +0.83dB in PSNR compared with ID. 2. It significantly reduces training time and boosts rendering performance. After obtaining high-quality novel view RGB images, the Enhanced Depth-Prior Training offers a robust geometric framework. This framework assists the Gaussian representations in aligning more accurately with large planar surfaces, improving overall geometric consistency, as shown in Fig.~\ref{fig:ablation_vis}. However, fine details can not be obtained by this process, and the vehicle in the lower-left corner is inaccurately estimated. So, Batch-Level Geometric Training is designed to refine and correct the detailed geometry of the scene, as it can be seen from the third row of depth in Fig.~\ref{fig:ablation_vis}. Moreover, Depth-Prior Training provides a smooth transition, enabling better convergence of Batch-Level Geometric Training process. As in the Tab.~\ref{tab:progressive}, ID 4 achieves improvement +0.66 and shorter training time compared with ID. 5.

\noindent \textbf{Batch Size and GPU Number.} Tab.~\ref{tab: Batch_ab} shows that, with a fixed total number of training steps and similar model storage, larger batch sizes generally improve model performance. For instance, Batch Size 16 achieves improvements of +0.7, +0.06, and +0.03 over Batch Size 2. This improvement is primarily due to the aggregation of multi-view gradients, which enables PH\(^2\)-3D to consider multi-view constraints, rather than overfitting to a single view. 


\begin{table}[t]
 \centering
   \caption{\textbf{Comparison with SOTA reconstruction methods on MatrixCity~\cite{li2023matrixcity} dataset.} $\uparrow$: higher is better, $\downarrow$: lower is better.
 }
 \vspace{-0.1in}
 \setlength{\tabcolsep}{9pt}
        \footnotesize
        \begin{tabular}{l cccc }
            \toprule
            Method &  PSNR$\uparrow$ & P$\uparrow$ & R$\uparrow$ & F1$\uparrow$   \\ \midrule
            NeuS  
             & 16.76 & FAIL & FAIL & FAIL\\
            Neuralangelo

             & 19.22 & 0.080 & 0.083 & 0.081 \\

            SuGaR

            & OOM & OOM & OOM & OOM \\

            GOF

            & 17.42 & FAIL & FAIL & FAIL \\ 

            2DGS

            & 21.35 & 0.207 & 0.390 & 0.270\\

            CityGS 

            & 27.46 & 0.362 & 0.637 &0.462\\
        
            CityGS-V2

            &27.23 &0.441 &0.752 &0.556 \\
            \midrule
            \textbf{CityGS-\(\mathcal{X}\) (Ours)}  
            & \textbf{27.58} & \textbf{0.444}  & \textbf{0.840}  & \textbf{0.581}
            \\
            \bottomrule
        \end{tabular}
    \label{tab: compare_reconstruction}
\centering
\end{table}

\begin{table}[t]
\centering
\footnotesize
\caption{\textbf{Ablations of Batch Size (B.S.) and GPU number on Rubble~\cite{turki2022mega}.} Besides the rendering performance, we present the final model's strage (MB) and singe GPU's Memory consumption (GB) during evaluation.}
\begin{tabular}{cccccc}
 \toprule
\multicolumn{1}{l}{B.S. / GPU} & PSNR$\uparrow$ & SSIM$\uparrow$ & LPIPS$\downarrow$ & Storge  & Mem\\ \midrule
 2 / 2 &25.58 &0.787 &0.244    & $1540\times 2$  & 14.76\\
 4 / 4 &25.70 &0.812 &0.234 & $775\times 4$ & 14.31\\  
 8 / 4 &26.15 &0.823 &0.210 & $788\times 4$ & 20.16\\  
 16 / 8 &\textbf{26.25} &\textbf{0.825} &\textbf{0.210} &$ 379\times 8$ &17.57  \\ \bottomrule
\end{tabular}
\label{tab: Batch_ab}
\vspace{-0.1in}
\end{table}

\begin{table}[t]
\centering
\footnotesize
\caption{\textbf{Ablations of Batch-Level Consistent Progressive Training on Building~\cite{turki2022mega}.}}
\vspace{-0.1in}
\setlength{\tabcolsep}{4pt}
\begin{tabular}{cccccc}
\toprule
\multirow{2}{*}{ID} &
  \multirow{2}{*}{\begin{tabular}[c]{@{}c@{}}Batch-Level\\
  RGB Training\end{tabular}} &
  \multirow{2}{*}{\begin{tabular}[c]{@{}c@{}}Depth-Prior \\Training\end{tabular}} &
  \multirow{2}{*}{\begin{tabular}[c]{@{}c@{}}Batch-Level\\Geometric Training\end{tabular}}  &
  \multirow{2}{*}{\begin{tabular}[c]{@{}c@{}}PSNR\end{tabular}} &
  \multirow{2}{*}{\begin{tabular}[c]{@{}c@{}}Time\end{tabular}} \\
  &   &   &   &   &    \\
\midrule
1 & \color[HTML]{CB0000}\textbf{\XSolidBrush} & {\color[HTML]{CB0000}\XSolidBrush} & {\color[HTML]{CB0000}\XSolidBrush} &22.41 & 4h21m \\
2 & \color[HTML]{009901}\textbf{\Checkmark} & {\color[HTML]{CB0000}\XSolidBrush} & {\color[HTML]{CB0000}\XSolidBrush} &23.24 & 1h48m \\
3 & \color[HTML]{009901}\textbf{\Checkmark} & \color[HTML]{009901}\textbf{\Checkmark} & {\color[HTML]{CB0000}\XSolidBrush} &22.46 &2h10m  \\
4 & \color[HTML]{009901}\textbf{\Checkmark} & \color[HTML]{009901}\textbf{\Checkmark} & \color[HTML]{009901}\textbf{\Checkmark} &22.76 &3h04m \\
5 & \color[HTML]{009901}\textbf{\Checkmark} & \color[HTML]{CB0000}\textbf{\XSolidBrush} & \color[HTML]{009901}\textbf{\Checkmark} &22.10 &3h11m \\
\bottomrule
\end{tabular}

\label{tab:progressive}
\vspace{-0.2in}
\end{table}
\section{Conclusions}
In this paper, we propose CityGS-\(\mathcal{X}\), a scalable architecture build on a novel parallelized hybrid hierarchical 3D representation (PH$^2$-3D). Unlike conventional methods that rely on a partition-and-merge strategy, our approach leverages PH$^2$-3D to achieve efficient load balancing while reducing the number of Gaussians needed for rendering, significantly enhancing efficiency. Furthermore, we apply Batch-Level Consistent Progressive Training to further boost appearance quality and geometric accuracy. Comprehensive experiments demonstrate that CityGS-\(\mathcal{X}\) achieves significant improvement of training speed and reconstruction quality of both appearance and geometry.

{
    \small
    \bibliographystyle{ieeenat_fullname}
    \bibliography{main}
}

\clearpage
\setcounter{page}{1}
\maketitlesupplementary

\section*{Implementation Details}
Given that Mill-19~\cite{turki2022mega} and UrbanScene3D~\cite{lin2022capturing} consist of thousands of high-resolution images, we adhere to the methodology outlined in previous works~\cite{turki2022mega} by downsampling the images by a factor of 4 for both training and validation. 
For evaluation, we adopt the configuration from Momentum-GS~\cite{fan2024momentumgsmomentumgaussianselfdistillation}, which excludes color correction when computing the metrics. 
Regarding the depth-prior filter, we set the threshold \(\tau_{d}\) to 1. 
In the training process for the Rubble, Building, Residence, and Sci-Art datasets, we define the total number of training steps as 100,000. The anchor growing process is maintained until the 50,000th step. The Step 2 Depth-Prior loss is introduced at the 10,000th iteration, with its weight progressively decreasing from 1 to 0 as training advances. The Step 3 Batch-Level Geometric Training is initiated at the 30,000th iteration, during which the weight of this loss incrementally rises from 0 to 0.2 throughout the training duration.

Given that MatrixCity~\cite{li2023matrixcity} comprises over 5000 images, we conducted training for 150,000 iterations. Following the approach of CityGS-V2~\cite{liu2024citygaussianv2}, we downsampled the images by a factor of 1.2 for training purposes. Owing to the scene's relatively simple geometric structure, the monocular depth estimator method performs with greater accuracy in this context. Consequently, we introduced the Enhanced Depth-Prior Training (Step 2) at the 10,000th iteration and implemented the Batch-Level Geometric Training (Step 3) at the 100,000th iteration.

For mesh reconstruction, we render both visual images and depth maps from multiple viewpoints. These rendered outputs are subsequently fused into a projected truncated signed distance function (TSDF) volume ~\cite{zeng20173dmatch}, ultimately generating high-quality 3D surface meshes and point clouds.
We set the voxel size of 0.01 and SDF truncation of 0.04 in MatrixCity, Residence, and Sci-Art, while 0.001, 0.004 in Rubble and Building datasets. 

\begin{figure}[htbp]
    \centering
    \includegraphics[width=1\linewidth]{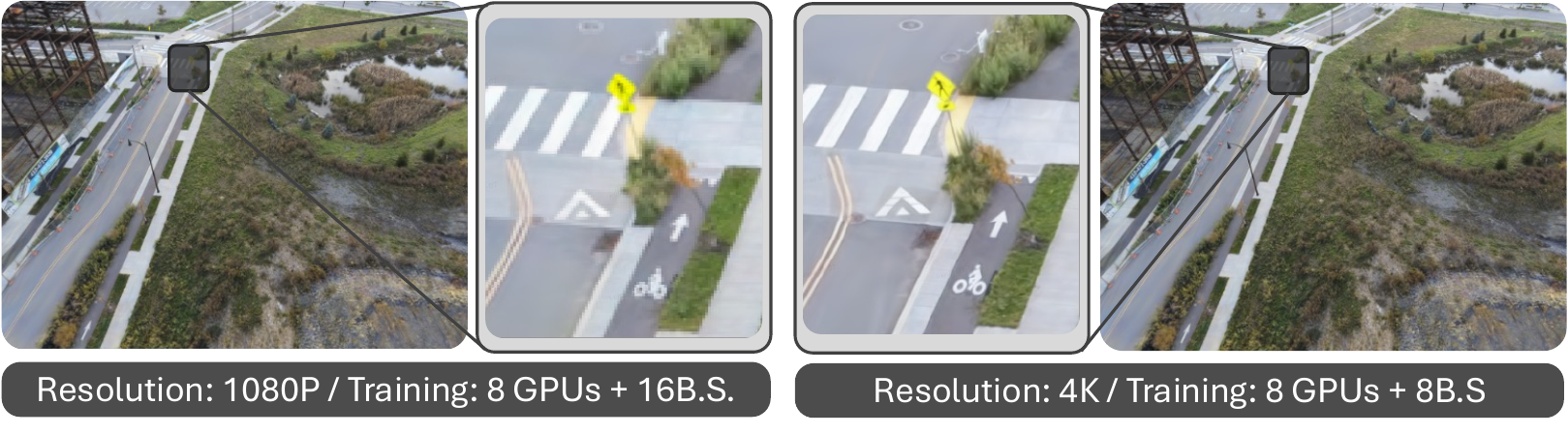}
    \caption{Visual comparison between 1080P and 4K training with our method.}
    \label{fig:4k}
\end{figure}

\begin{figure}[t]
    \centering
    \includegraphics[width=1\linewidth]{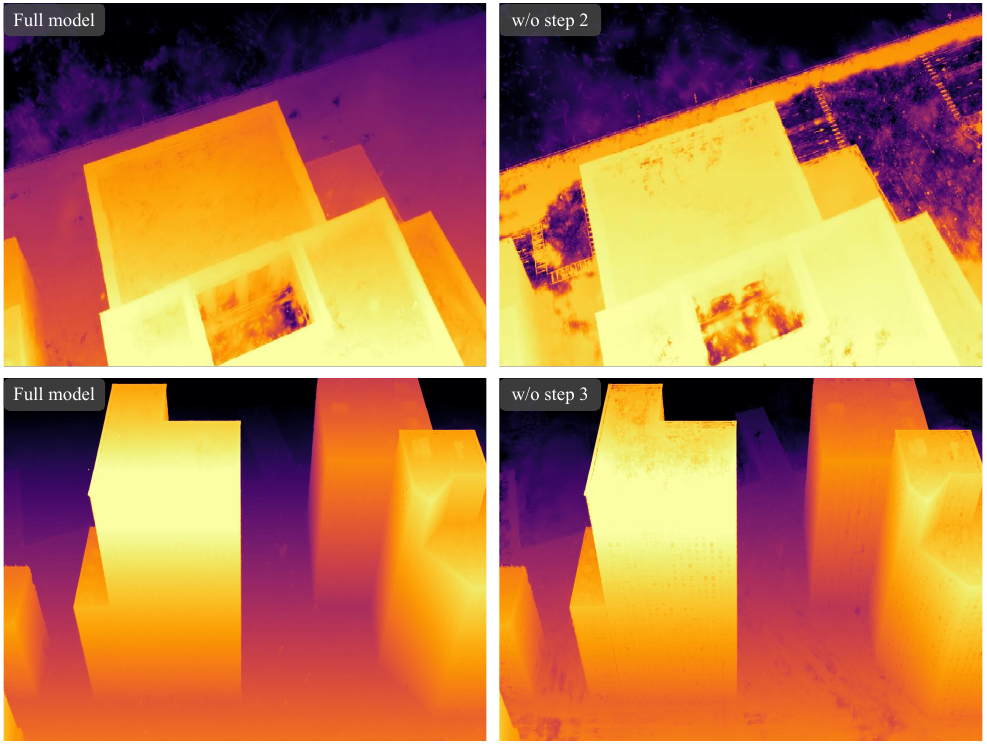}
    \caption{Training ablations for MatrixCity.}
    \label{fig:Matrix}
\end{figure}

\section*{4K Training and Rendering.}
To demonstrate that our method maintains high-quality reconstruction even at higher resolutions, we compare training results between 1080P and 4K settings, as shown in Fig.~\ref{fig:4k}. The 1080P model is trained with 8 GPUs and a batch size of 16, while the 4K model uses the same number of GPUs but a batch size of 8 due to increased memory consumption.
The qualitative comparison shows that our method also achieves consistently high reconstruction quality in 4K settings. The 4K model preserves fine-grained details, such as sharp edges, clear road markings, and accurate object boundaries, without introducing artifacts or degradation. This indicates that our approach effectively scales to higher resolutions, ensuring robustness and reliability in large-scale urban scene reconstruction.

\begin{table*}[t]
\small
 \centering
 \caption{\textbf{Training Resources Consumption on Mill19~\cite{turki2022mega} dataset and UrbanScene3D~\cite{lin2022capturing} dataset}.
 We present the allocated memory (GB) during evaluation. For 3DGS-based methods, .
 }
 \label{table:urban3d_quantitative_time}
 \begin{tabular}{l  l l l l  l l l l  l l l l  l l l }
 \toprule
 
 \multirow{2}{*}{Models} &
 \multicolumn{2}{c}{Building} &
 \multicolumn{2}{c}{Rubble} & 
 \multicolumn{2}{c}{Residence} & 
 \multicolumn{2}{c}{Sci-Art} \\
 
 \cmidrule(r){2-3} \cmidrule(r){4-5} \cmidrule(r){6-7} \cmidrule(r){8-9}

 & \multirow{1}{*}{Time $\downarrow$}
 & \multirow{1}{*}{Mem $\downarrow$} 
 
 & \multirow{1}{*}{Time $\downarrow$}
 & \multirow{1}{*}{Mem $\downarrow$} 
 
 & \multirow{1}{*}{Time $\downarrow$}
 & \multirow{1}{*}{Mem $\downarrow$} 

 & \multirow{1}{*}{Time $\downarrow$}
 & \multirow{1}{*}{Mem $\downarrow$} \\

 \midrule

 Mega-NeRF~\cite{turki2022mega}
 & 19:49  & 5.84 
 & 30:48  & 5.88 
 & 27:20  & 5.99 
 & 27:39  & 5.97 \\

 Switch-NeRF~\cite{zhenxing2022switch}
 & 24:46  & 5.84  
 & 38:30  & 5.87  
 & 35:11  & 5.94  
 & 34:34  & 5.92  \\

 $\text{3DGS}$~\cite{kerbl3Dgaussians}
 & 21:37  & 4.62  
 & 18:40  & 2.18 
 & 23:13  & 3.23  
 & 21:33  & 1.61  \\

 $\text{VastGS}^{\dagger}$~\cite{lin2024vastgaussian}
 & 03:26  & 3.07 
 & 02:30  & 2.74 
 & 03:12  & 3.67 
 & 02:33  & 3.54 \\

 DOGS~\cite{chen2024dogaussian}
 & 03:51  & 3.39 
 & 02:25  & 2.54 
 & 04:33  & 6.11 
 & 04:23  & 3.53  \\

 
 \midrule
  CityGS-\(\mathcal{X}\)
 & \textbf{03:00}  & \textbf{2.00}
 & \textbf{02:15}  & \textbf{2.29} 
 & \textbf{02:40}  & \textbf{2.61} 
 & \textbf{03:30}  & \textbf{1.40} 
 \\ \bottomrule
\end{tabular}
\end{table*}

\begin{figure*}[t]
    \centering
    \includegraphics[width=1\linewidth]{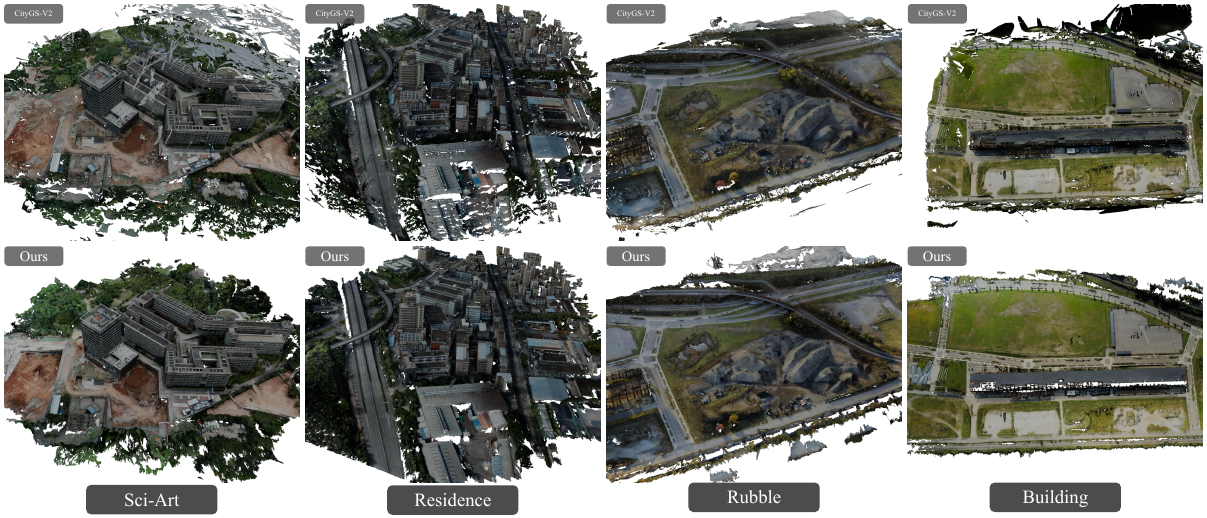}
    \caption{Qualitative comparison between CityGS-V2~\cite{liu2024citygaussianv2} and ours in generated textured meshes.}
    \label{fig:suppl_texture}
\end{figure*}

\begin{figure*}[t]
    \centering
    \includegraphics[width=1\linewidth]{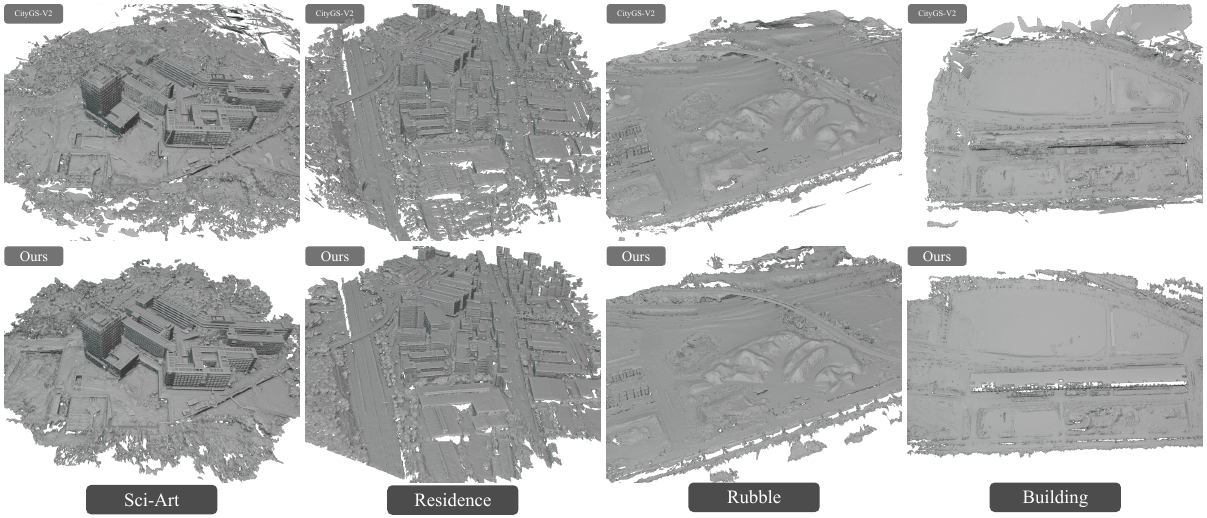}
    \caption{Qualitative comparison between CityGS-V2~\cite{liu2024citygaussianv2} and ours in generated meshes.}
    \label{fig:suppl_mesh}
\end{figure*}

\begin{figure*}[t]
    \centering
    \includegraphics[width=1\linewidth]{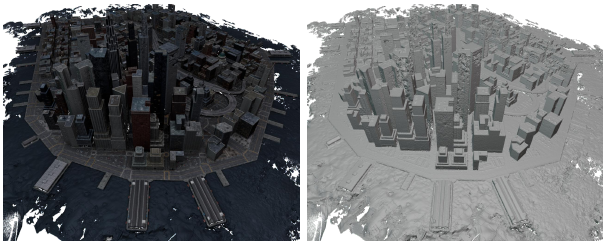}
    \caption{Qualitative results of our generated meshes on MatrixCity~\cite{li2023matrixcity} dataset.}
    \label{fig:suppl_matrix}
\end{figure*}

\section*{Overall Mesh Visualization}
In Fig.~\ref{fig:suppl_texture} and Fig.~\ref{fig:suppl_mesh}, we visualize textured meshes and meshes of our method and CityGSV2~\cite{liu2024citygaussianv2} on Sci-Art, Residence, Rubble, Building. Our mesh clearly exhibits superior geometric structure, demonstrating enhanced accuracy and detail compared to CityGSV2. In the Sci-Art, our mesh has fewer floaters, while in the bottom of the Residence and Rubble we have fewer holes. In the Building datasets, for the central building, we preserve its texture while maintaining accurate structural integrity, a capability that CityGSv2 fails to achieve.

\section*{Training Strategy Ablation on MatrixCity}
In Fig.~\ref{fig:Matrix}, in the first row we directly add the Batch-Level Geometric Training (Step 3) without the transition of Step 2, and the large plane of the scene on the ground is hard to converge. In the second row, without Step 3, some geometric details on the ground are not accurate. Therefore, both Step 2 and Step 3 are necessary for optimizing geometry.

\section*{Training Resources Consumption}
Tab.~\ref{table:urban3d_quantitative_time} presents the training resource consumption of various methods on the Mill19 and UrbanScene3D datasets. It reports the training time (Time ↓) and memory usage in GB (Mem ↓) across four different scenes: Building, Rubble, Residence, and Sci-Art.

Among all methods, CityGS-X achieves the fastest training time and lowest memory consumption across all scenes. Specifically, it requires only 3:00 minutes and 2.00 GB for the Building scene, 2:15 minutes and 2.29 GB for Rubble, 2:40 minutes and 2.61 GB for Residence, and 3:30 minutes and 1.40 GB for Sci-Art. In contrast, Mega-NeRF and Switch-NeRF take significantly longer training times (e.g., up to 38:30 minutes for Rubble) and higher memory usage (~5.9 GB). 3DGS, VastGS, and DOGS show moderate resource consumption, but CityGS-X consistently outperforms them in both speed and efficiency.
These results highlight the superior efficiency of CityGS-X, making it a more practical choice for large-scale scene reconstruction with limited computational resources.

\end{document}